\newcommand{\zhou}[1]{{\color{black}{#1}}}
\documentclass[conference]{IEEEtran}
\IEEEoverridecommandlockouts

\usepackage{cite}
\usepackage{amsmath,amssymb,amsfonts}
\usepackage{algorithmic}
\usepackage{graphicx}
\usepackage{textcomp}
\usepackage{xcolor}
\usepackage{booktabs,multirow}
\usepackage{natbib} 
\usepackage{graphicx}
\usepackage{cleveref}
\usepackage{caption}

\crefalias{paragraph}{section} 

\def\BibTeX{{\rm B\kern-.05em{\sc i\kern-.025em b}\kern-.08em
    T\kern-.1667em\lower.7ex\hbox{E}\kern-.125emX}}
\begin{document}

\title{AbICL: In-Context Learning for Antigen-Specific Antibody Affinity Ranking
\thanks{$^{\dagger}$Corresponding author}%
}


\author{\IEEEauthorblockN{Zhiyuan Chen$^{\dagger}$}
\IEEEauthorblockA{
\textit{Henlius}\\
Shanghai, China \\
Zhiyuan\_Chen@henlius.com}
\and
\IEEEauthorblockN{Jing Hu$^{\dagger}$}
\IEEEauthorblockA{
\textit{Henlius}\\
Shanghai, China \\
Jing\_Hu1@henlius.com}
\and
\IEEEauthorblockN{Junzhe Wang}
\IEEEauthorblockA{
\textit{Henlius}\\
Shanghai, China \\
Junzhe\_Wang@henlius.com}
\and
\IEEEauthorblockN{Yueyang Huang}
\IEEEauthorblockA{
\textit{Henlius}\\
Shanghai, China \\
Yueyang\_Huang@henlius.com}
\and
\IEEEauthorblockN{Xinyi Yang}
\IEEEauthorblockA{
\textit{Henlius}\\
Shanghai, China \\
Connie\_Yang@henlius.com}
\and
\IEEEauthorblockN{Zhaoyang Wang}
\IEEEauthorblockA{
\textit{Henlius}\\
Shanghai, China \\
John\_Wang1@henlius.com}
\and
\IEEEauthorblockN{Feng Zhu}
\IEEEauthorblockA{
\textit{Henlius}\\
Shanghai, China \\
Shawn\_Zhu@henlius.com}
}

\maketitle

\begin{abstract}
Accurate ranking of antibody candidates according to their binding affinity is essential for therapeutic antibody discovery. However, existing methods treat affinity comparisons independently and ignore the contextual information encoded in other labeled comparisons, limiting their ability to capture antigen-specific binding landscapes. For many target antigens, a small number of experimentally characterized affinity comparisons are often available. An important question is whether the model can exploit these existing comparisons to infer antigen-specific ranking patterns that facilitate subsequent affinity ranking. This form of learning from labeled demonstrations closely resembles the paradigm of In-Context Learning, motivating us to revisit antibody affinity ranking from an ICL perspective. To this end, we propose AbICL, an ICL framework for antigen-specific antibody affinity ranking. AbICL combines a pretrained structural encoder with a context ranking head and is trained with an episodic meta-training strategy that enables the model to leverage support demonstrations for test-time adaptation without gradient updates. Experiments on the AbRank benchmark demonstrate that AbICL consistently outperforms existing ranking baselines across almost all data splits and evaluation benchmarks. Further analysis shows that the value of contextual demonstrations depends on how well they match the target inference task, and becomes increasingly pronounced under distribution shift and fine-grained affinity discrimination.  These findings highlight the potential of ICL as an effective paradigm for antigen-specific antibody affinity ranking, particularly in challenging settings where a single global ranking function is insufficient.
\end{abstract}


\section{Introduction}
Predicting antibody-antigen binding affinity is central to therapeutic antibody discovery, yet direct affinity prediction remains challenging because affinity measurements are often noisy, assay-dependent, and frequently censored. Consequently, affinity ranking, which prioritizes candidates according to their relative binding strengths rather than absolute affinity values, has emerged as a robust alternative.

Existing approaches for antibody affinity prediction can be broadly categorized into three paradigms: affinity regression methods \cite{10.1093/bioinformatics/btag109,ullanat2026learning,10.1093/bioinformatics/btae579,10.1093/bioinformatics/btag109,yuan2023dg,bandara2025deep,hummer2025investigating,subedy2026se3bind,chen2025predicting,cai2024pretrainable} , ranking-based methods \cite{liu2025abrank,xu2026ablwr}, and indirect affinity scoring methods based on pretrained foundation models \cite{lin2023evolutionary, rives2021biological, candido2026language,abramson2024accurate,krishna2024generalized,wohlwend2025boltz,passaro2025boltz,chai2024chai,team2025zero,bytedance2025protenix,xu2026deeprank, ruffolo2023fast}. Affinity regression methods directly predict experimental binding affinities but are often affected by noisy and heterogeneous measurements. Ranking-based methods instead learn relative binding preferences, providing a more robust objective, yet typically rely on a context-independent ranking function that treats each comparison as an isolated prediction problem. Indirect affinity scoring methods estimate binding quality using sequence likelihoods, structural confidence, or pretrained protein representations rather than explicitly modeling antibody-antigen affinity. Despite their differences, these approaches all rely on fixed scoring functions or representations learned during training, limiting their ability to adapt to previously unseen antigens and distribution shifts.

However, antibody affinity ranking is inherently antigen-specific, as the relative preference between antibody candidates is determined by the binding landscape of the target antigen rather than a universal ranking function. This raises a natural question: rather than relying solely on a fixed ranking function learned during training, can predictions benefit from using by using antigen-specific evidence available? One possible source of such evidence is previously characterized affinity comparisons, which may provide useful information for inferring antigen-specific ranking patterns and facilitating subsequent ranking decisions. This way of conditioning predictions on labeled examples closely resembles the paradigm of ICL, where models adapt their predictions according to contextual demonstrations without updating model parameters. Motivated by this connection, we revisit antibody affinity ranking from an ICL perspective.

To this end, we propose AbICL, an ICL framework for antigen-specific antibody affinity ranking. AbICL combines a pretrained structural encoder with a context ranking head that jointly reasons over support demonstrations and query pairs. The model is trained via episodic meta-training to mimic the support-query scenarios encountered at inference time, enabling test-time adaptation through contextual demonstrations alone and without gradient updates.

Extensive experiments on the AbRank benchmark demonstrate that AbICL consistently outperforms existing ranking and regression baselines across all data splits and evaluation benchmarks. Further analysis shows that the benefits of contextual demonstrations become increasingly pronounced under distribution shift and fine-grained affinity discrimination, and are greatest when demonstrations closely match the target inference task.

Our contributions are summarized as follows:

\begin{itemize}
    \item We introduce the first ICL framework for antigen-specific antibody affinity ranking, enabling test-time adaptation through labeled affinity comparisons.

    \item We propose a context-dependent ranking framework that combines a pretrained structural encoder with a Context Ranking Head trained via episodic meta-training, enabling the model to effectively leverage contextual demonstrations for antigen-specific affinity ranking.

    \item We achieve state-of-the-art performance on the AbRank benchmark and provide comprehensive analyses showing that the gains of AbICL arise from effective in-context adaptation rather than increased model capacity. Our results further reveal that contextual demonstrations become increasingly valuable under distribution shift and fine-grained affinity discrimination.
\end{itemize}

\section{RELATED WORK}

\subsection{Antibody Affinity Prediction}
\paragraph{Affinity Regression Methods}
Affinity regression methods formulate antibody-antigen affinity modeling as a regression problem, aiming to directly predict either absolute binding affinity or affinity changes induced by mutations.Sequence-based approaches learn regression models from antibody and antigen sequences to predict binding affinity  \cite{10.1093/bioinformatics/btag109,ullanat2026learning,10.1093/bioinformatics/btae579,10.1093/bioinformatics/btag109,yuan2023dg} or mutation-induced affinity changes \cite{10.1093/bib/bbae304, liu2025sequence}. Complementary to sequence-based methods, structure-based approaches leverage experimentally determined or computationally predicted complex structures to explicitly model intermolecular interactions for affinity prediction \cite{bandara2025deep, hummer2025investigating, subedy2026se3bind, chen2025predicting, cai2024pretrainable}. Although these methods have achieved promising performance, they rely on supervised regression of continuous affinity values or affinity changes, which are often noisy, assay-dependent, and difficult to compare across experimental conditions.

\paragraph{Ranking-based Methods}
Motivated by the limitations of affinity regression, ranking-based approaches have recently emerged as an alternative paradigm for antibody-antigen affinity prediction by directly optimizing the relative ordering of antibody candidates rather than their absolute affinity values. Existing methods can be broadly categorized into pairwise and listwise ranking approaches. Pairwise methods learn relative preferences between pairs of antibody-antigen complexes and optimize pairwise ranking objectives to determine which candidate exhibits stronger binding affinity \cite{liu2025abrank}. In contrast, listwise methods jointly optimize the ordering of a set of antibody candidates for a given antigen, directly modeling the ranking of multiple candidates within a screening pool \cite{xu2026ablwr}. Despite their encouraging performance, existing ranking-based methods treat each ranking instance independently and rely on a fixed ranking function learned during training, without leveraging contextual information from previously characterized affinity comparisons.

\paragraph{Indirect Affinity Scoring Methods}
Recent advances in protein foundation models have inspired alternative strategies for antibody-antigen affinity estimation without task-specific affinity training. Sequence-based approaches leverage pretrained protein or antibody language models to score candidate sequences using likelihood or pseudo-likelihood, where sequence naturalness serves as a proxy for binding affinity \cite{lin2023evolutionary, rives2021biological, candido2026language}. Structure-based approaches instead \zhou{estimating} affinity from predicted antibody-antigen complex structures generated by foundation models such as AlphaFold3 \cite{abramson2024accurate}, RoseTTAFoldAA \cite{krishna2024generalized}, Boltz \cite{wohlwend2025boltz, passaro2025boltz}, Chai \cite{chai2024chai, team2025zero}, Protenix \cite{bytedance2025protenix}, or antibody-specific structure prediction models \cite{xu2026deeprank, ruffolo2023fast}, typically using structural confidence or interface quality metrics as indirect indicators of binding strength. Although these methods exhibit strong generalization in protein sequence modeling and structure prediction, they are not explicitly trained on antibody-antigen affinity data. Sequence-based scoring often overlooks antigen-specific binding interactions, while structure-based scoring relies on structural confidence or interface quality, which may not faithfully reflect binding affinity due to the complexity of antibody-antigen interactions and the flexibility of antibody CDRs.

\subsection{Episodic Meta-learning}
Meta-learning, or learning to learn, aims to facilitate rapid adaptation to novel tasks with minimal supervision. This is typically achieved by training across a collection of tasks that mimic the few-shot test-time scenario. Existing literature predominantly falls into three categories. Metric-based methods, such as Matching Networks \cite{vinyals2016matching} and Prototypical Networks \cite{snell2017prototypical}, focus on learning a shared embedding space where classification is performed via distance metrics. Optimization-based methods, exemplified by MAML \cite{finn2017model} and Reptile \cite{nichol2018first}, seek to learn a task-agnostic parameter initialization for fast gradient adaptation. Additionally, model-based approaches, such as SNAIL \cite{mishra2017simple} and MANN \cite{santoro2016meta}, utilize internal memory or attention mechanisms to rapidly integrate support information.

To bridge the gap between training and evaluation, the episodic training paradigm organizes the process into a series of episodes \cite{vinyals2016matching}. Each episode typically comprises a support set for task adaptation and a query set for performance evaluation \cite{snell2017prototypical}. By simulating few-shot constraints during training, this paradigm forces the model to learn task-level representations rather than memorizing class-specific labels, thereby enhancing generalization to unseen tasks \cite{finn2017model}.

\subsection{In-Context Learning for Biological Prediction}
ICL has emerged as a fundamental capability of large language models \cite{NEURIPS2020_1457c0d6}, enabling models to adapt to new tasks by conditioning on a small set of labeled demonstrations without updating model parameters. Among the many theoretical perspectives proposed to explain this capability, two influential views suggest that Transformers either implicitly perform gradient-based adaptation during the forward pass \cite{von2023transformers} or act as learned meta-optimizers that perform task-specific adaptation through contextual demonstrations \cite{dai2023can}. Beyond natural language processing, recent studies have extended ICL to molecular property prediction \cite{fifty2023context, 3737916.3740177} and protein fitness prediction \cite{beck2025metalic}, where contextual demonstrations are incorporated to improve prediction performance. To the best of our knowledge, ICL has not been explored for antibody-antigen affinity prediction, where antigen-specific binding preferences present unique challenges that cannot be adequately addressed by a single global prediction function.

\section{METHODOLOGIES}

\begin{figure*}[t]
\centering
\includegraphics[width=1.0\linewidth]{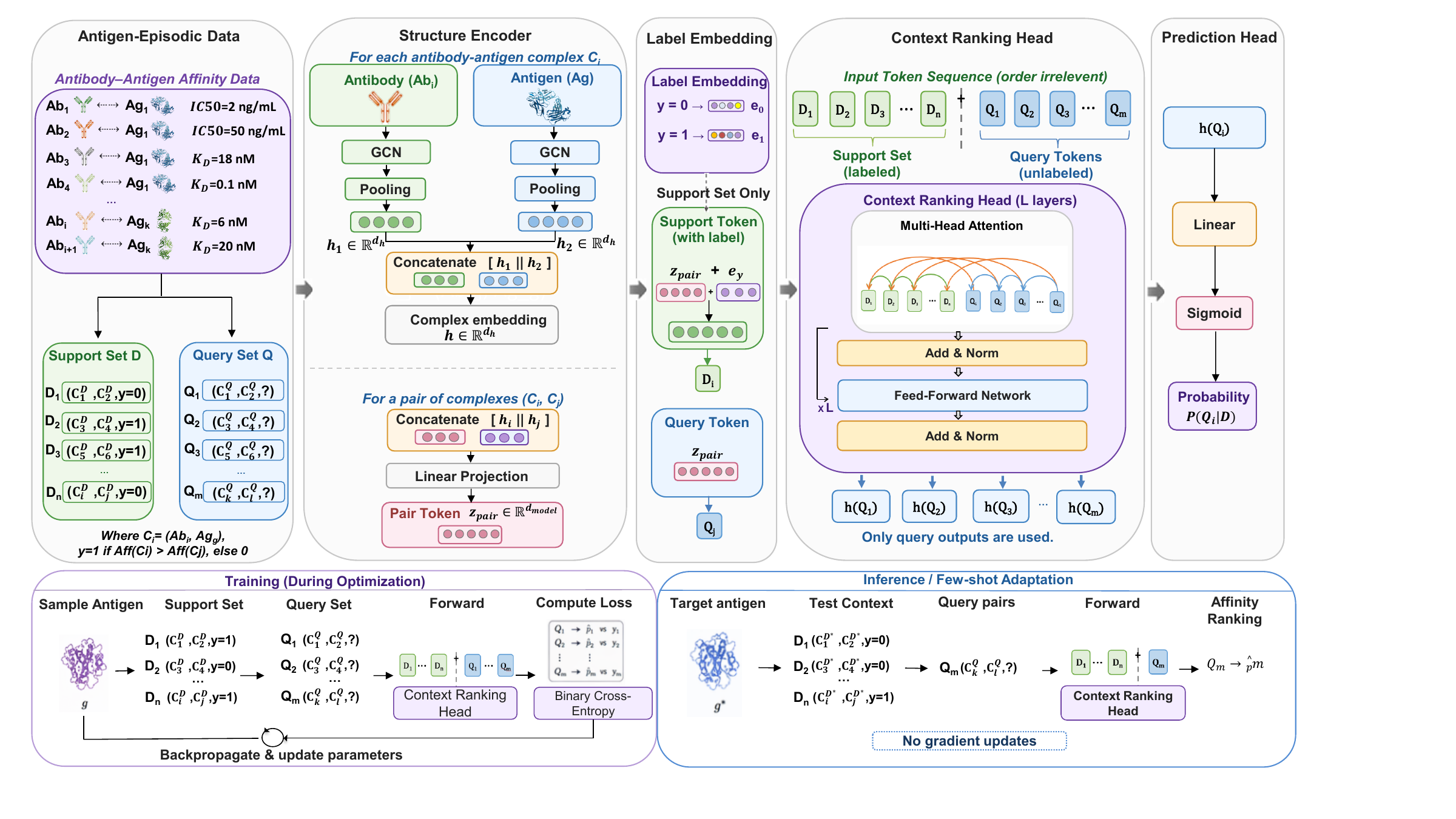}
\caption{The framework of AbICL.
}
\label{fig:framework}
\end{figure*}

\subsection{Problem Formulation}

Given a target antigen $Ag$, let
\(
\mathcal{A}=\{Ab_1,Ab_2,\ldots\}
\)
denote a set of candidate antibodies. Each antibody together with the target antigen forms an antibody--antigen complex
\(
C_i=(Ab_i,Ag).
\)

We formulate antibody--antigen affinity ranking as a pairwise classification problem. Given a pair of complexes
\(
(C_i,C_j)
\)
sharing the same antigen, the model predicts which complex exhibits stronger binding affinity:
$P(C_i \succ C_j)$. The ground-truth label is defined as

\begin{equation}
y=
\begin{cases}
1, & \mathrm{Aff}(C_i)>\mathrm{Aff}(C_j),\\
0, & \text{otherwise}.
\end{cases}
\end{equation}



\subsection{Episode Construction}
\label{sec:episodes}

AbICL is trained using an antigen-episodic meta-training strategy, where each training episode is constructed from a single target antigen \(Ag\). For a given antigen, let
\(
\mathcal{D}_{Ag}
\)
denote the set of all labeled pairwise ranking instances associated with that antigen, where each instance consists of a pair of complexes and its binary ranking label:
$
\mathcal{D}_{Ag}
=
\{
(C_i,C_j,y)
\}$.

At each training step, the instances in
\(
\mathcal{D}_{Ag}
\)
are randomly partitioned into

\begin{itemize}
    \item \textbf{Support set}
    $    
    \mathcal{D}
    =
    \{
    (C_i,C_j,y)
    \}_{i=1}^{K},
    $
    containing \(K\) labeled demonstrations, where
    \(K\sim\mathrm{Uniform}(0,K_{\max})\).

    \item \textbf{Query set}
    $
    \mathcal{Q}
    =
    \{
    (C_i,C_j)
    \}_{i=1}^{Q},
    $
    consisting of the remaining comparison instances for the same antigen, where \(Q\le Q_{\max}\).
\end{itemize}

Since all query instances in an episode correspond to the same target antigen, they naturally share a common support set and can therefore be evaluated simultaneously.

During training, the support size \(K\) is randomly sampled to expose the model to varying amounts of contextual demonstrations. Additional details on episode construction and sampling procedures are provided in \Cref{support_build}.

\subsection{Model Architecture}

\subsubsection{Structure Encoder}
Given an antibody--antigen complex
\(
C=(Ab,Ag)
\),
we represent the antibody and antigen as two molecular graphs,
\(G_{Ab}\) and \(G_{Ag}\), respectively.
Following WALLE-Affinity \cite{liu2025abrank}, the antibody graph and antigen graph are encoded by two separate graph convolutional networks (GCNs), and their graph-level representations are obtained via global mean pooling:

\begin{equation}
\begin{aligned}
h_{Ab} &= \mathrm{pool}\!\left(\mathrm{GCN}_{Ab}(G_{Ab})\right),\\
h_{Ag} &= \mathrm{pool}\!\left(\mathrm{GCN}_{Ag}(G_{Ag})\right),\\
h &= [h_{Ab};\,h_{Ag}]
\in\mathbb{R}^{d_{\mathrm{enc}}}.
\end{aligned}
\end{equation}

Given a pair of complexes \(
(C_i,C_j)
\),
where
\(
C_i=(Ab_i,Ag)
\)
and
\(
C_j=(Ab_j,Ag)
\),  we independently encode them into representations \(h_i\) and \(h_j\), which are further fused into a pair representation:

\begin{equation}
z_{\mathrm{pair}}
=
W
\left[
h_i;
h_j
\right]
\in
\mathbb{R}^{d_{\mathrm{model}}},
\end{equation}

where
\(
[\,;\,]
\)
denotes concatenation.

\subsubsection{Label Embedding}

To enable ICL, each support pair representation is augmented with its ground-truth ranking label through a learnable embedding:

\begin{equation}
\tilde{z}^{(s)}_i
=
z^{(s)}_{\mathrm{pair},i}
+
\mathrm{LabelEmbed}(y_i),
\qquad
y_i\in\{0,1\},
\end{equation}

where
\(
\mathrm{LabelEmbed}:\{0,1\}\rightarrow\mathbb{R}^{d_{\mathrm{model}}}
\)
is a learnable embedding table.  While we focus on binary ranking labels in this work, the proposed framework can be readily generalized to continuous labels by replacing the discrete embedding table with a learnable continuous label encoder (e.g., Gaussian radial basis expansion followed by an MLP \cite{schutt2017schnet}). 

Query pair representations do not receive label embeddings because their labels are unavailable during inference.

\subsubsection{Context Ranking Head}

The support and query pair representations are concatenated and jointly processed by a Pre-LayerNorm Transformer encoder \cite{vaswani2017attention}. Since the support demonstrations form an unordered set, their ordering should not affect the prediction. Therefore, no positional encoding is used, following the design of Set Transformer \cite{lee2019set}.

\begin{equation}
\begin{aligned}
&
[h^{(s)}_1,\ldots,h^{(s)}_K,\,
h^{(q)}_1,\ldots,h^{(q)}_Q]
\\
&=
\mathrm{TransformerEncoder}
\left(
[\tilde z^{(s)}_1,\ldots,\tilde z^{(s)}_K,\,
z^{(q)}_1,\ldots,z^{(q)}_Q]
\right).
\end{aligned}
\end{equation}

Only the query outputs are used for prediction:

\begin{equation}
P(C_i \succ C_j \mid \mathcal{D})
=
\sigma\!\left(
\mathrm{Linear}(h^{(q)}_j)
\right).
\end{equation}

The self-attention mechanism enables each query pair to attend to all labeled support pairs, allowing the model to adapt its affinity ranking predictions according to the contextual information encoded in the support demonstrations.

\subsection{Training Objective}

We minimize binary cross-entropy over all valid (non-padding) query pairs:
\begin{equation}
\begin{split}
    \mathcal{L} = -\frac{1}{|\mathcal{Q}|} & \sum_{j \in \mathcal{Q}} 
    \big[ y_j \log \sigma(\mathrm{logit}_j) \\
    & + (1 - y_j) \log (1 - \sigma(\mathrm{logit}_j)) \big]
\end{split}
\end{equation}

\section{Experiments}
\subsection{Experimental Setup}
\subsubsection{Dataset and baselines} We use the pairwise affinity ranking data from the AbRank benchmark \cite{liu2025abrank}. Each sample is a pair of Ab--Ag complexes sharing the same antigen, with a binary label indicating which complex has higher binding affinity. AbRank provides three training splits of increasing difficulty: 
\begin{itemize}
    \item Balanced: test complexes may be sequence-similar to training complexes.
    \item Hard Ab: each test antibody has less than $75\%$ sequence identity to any training antibody.
    \item Hard Ag: each test antigen has less than $75\%$ sequence identity to any training antigen, requiring generalization to entirely new antigenic targets.
\end{itemize}

All splits share the same two test benchmarks:
\begin{itemize}
    \item Unrelated Complex: evaluates cross-antigen generalization using pairs drawn from structurally unrelated antigens.
    \item Local Perturbation: evaluates sensitivity to fine-grained affinity differences among near-neighbor mutant variants of the same antigen.
\end{itemize}

We compare against WALLE-Affinity \cite{liu2025abrank}, ESM-2 + AntiBERTy \cite{ruffolo2021deciphering}, and Mint \cite{ullanat2026learning} as baselines.

\begin{table}[h]
\centering
\caption{Mean number of in-context demonstrations available per test antigen under each inference protocol.}
\label{tab:support-stats}
\resizebox{\columnwidth}{!}{%
\begin{tabular}{llcc}
\toprule
Benchmark & Train split & Training-context & Test-context \\
\midrule
\multirow{3}{*}{Generalization} & Balanced & 8.0 & 1.9 \\
                                & Hard Ab  & 8.0 & 1.9 \\
                                & Hard Ag  & 0.0 & 1.9 \\
\midrule
\multirow{3}{*}{Perturbation}   & Balanced & 6.9 & 1.1 \\
                                & Hard Ab  & 7.1 & 1.1 \\
                                & Hard Ag  & 0.0 & 1.1 \\
\bottomrule
\end{tabular}
}
\end{table}

\begin{figure*}[t]
\centering
\includegraphics[width=0.95\linewidth]{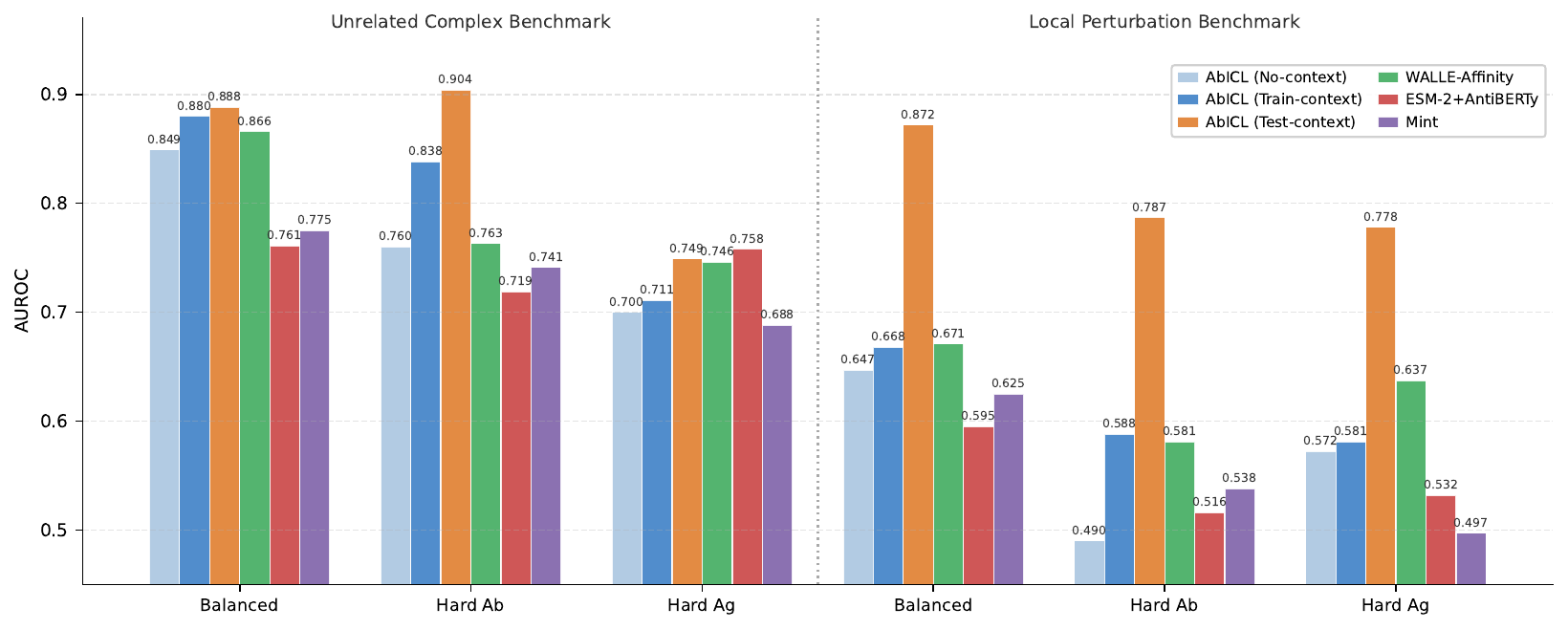}
\caption{AUROC performance on the AbRank benchmark.
}
\label{fig:overall}
\end{figure*}

\subsubsection{Training protocols}

Training proceeds in two stages.

\textbf{Stage 1} reproduces the WALLE-Affinity encoder~\cite{liu2025abrank}. The antibody and antigen GCN encoders (\(\mathrm{GCN}_{Ab}\) and \(\mathrm{GCN}_{Ag}\)) are trained using a margin ranking loss on the full training set to obtain a structural encoder pretrained for affinity discrimination. Training is performed for up to 60 epochs with early stopping based on validation AUROC (patience = 10, minimum improvement \(\delta=0.001\)).

\textbf{Stage 2} trains the Pair Representation layer, Label Embedding, and Context Ranking Head, while freezing the pretrained structural encoder from Stage~1. Training episodes are constructed as described in Section~\ref{sec:episodes}, with the support size randomly sampled from \([0,K_{\max}]\) and \(K_{\max}=8\). Stage 2 is trained for up to 20 epochs using the same early stopping strategy.

Both stages are optimized using AdamW with a one-cycle learning rate schedule, consisting of a peak learning rate of \(10^{-4}\), linear warmup over the first \(5\%\) of training steps, and cosine annealing thereafter. Gradients are clipped to unit norm, and the episode batch size is set to 8.

\subsubsection{Inference Protocols} \label{infer_proc}
We evaluate AbICL under three test-time inference protocols that differ in the source of in-context demonstrations. At test time, the model receives a set of labeled pair demonstrations as in-context examples, analogous to the support set used during training episodes: 
\begin{itemize}
    \item \textbf{No-context}: No demonstrations are provided; the model scores each query pair using only its own internal representations. This mirrors the standard evaluation setting of AbRank and serves as a baseline to isolate the contribution of the ICL mechanism.
    \item \textbf{Training-context}: Demonstrations are sampled from the training-set complexes of the same antigen (up to $K_{\max} = 8$ shots), representing a scenario where historical binding data for the target antigen are available from prior experiments.

    \item \textbf{Test-context}: Demonstrations are sampled from the test set itself for the same antigen, with the remaining test pairs used as queries (non-overlapping). This models a practical scenario where a small number of experimentally measured pairs for the target antigen are available at inference time, inspired by the CAMP \cite{fifty2023context} evaluation protocol.
\end{itemize}

\Cref{tab:support-stats} reports the mean number of demonstration pairs available per test antigen under each protocol. Under the training-context protocol, the Balanced split consistently provides 8 demonstration pairs per antigen episode, while the Hard Ag split yields no demonstrations at all, since no training antigen shares sufficient sequence identity with any test antigen. Under the test-context protocol, demonstrations are limited by the small number of test pairs available per antigen, resulting in roughly 1 demonstration per antigen on average. Additional details on inference context construction are provided in \Cref{context_build}.

\subsection{Main Results}

\Cref{fig:overall} shows AUROC on the AbRank benchmark under the three training splits (Balanced, Hard Ab, and Hard Ag) and two complementary evaluation benchmarks (Unrelated Complex and Local Perturbation). For each setting, we compare the three inference protocols introduced in \cref{infer_proc} (No-context, Training-context, and Test-context) to analyze the effect of contextual demonstrations under different inference scenarios.

\subsubsection{\textbf{Contextual demonstrations improve affinity ranking}}
Due to the unavailability of the warm-start initialization used in WALLE-Affinity, our reproduced structural encoder obtained in Stage1 is weaker than the original model. As a result, the No-context variant consistently underperforms the reported WALLE-Affinity baseline across most evaluation settings. Nevertheless, introducing contextual demonstrations markedly improves performance. With Training-context, AbICL already surpasses WALLE-Affinity on multiple benchmarks despite using the same reproduced encoder, while Test-context further achieves the best AUROC on nearly all evaluation settings. These results indicate that the performance gains originate from effective in-context adaptation rather than a stronger structural backbone.


\begin{figure*}[t]
\centering
\includegraphics[width=0.9\linewidth]{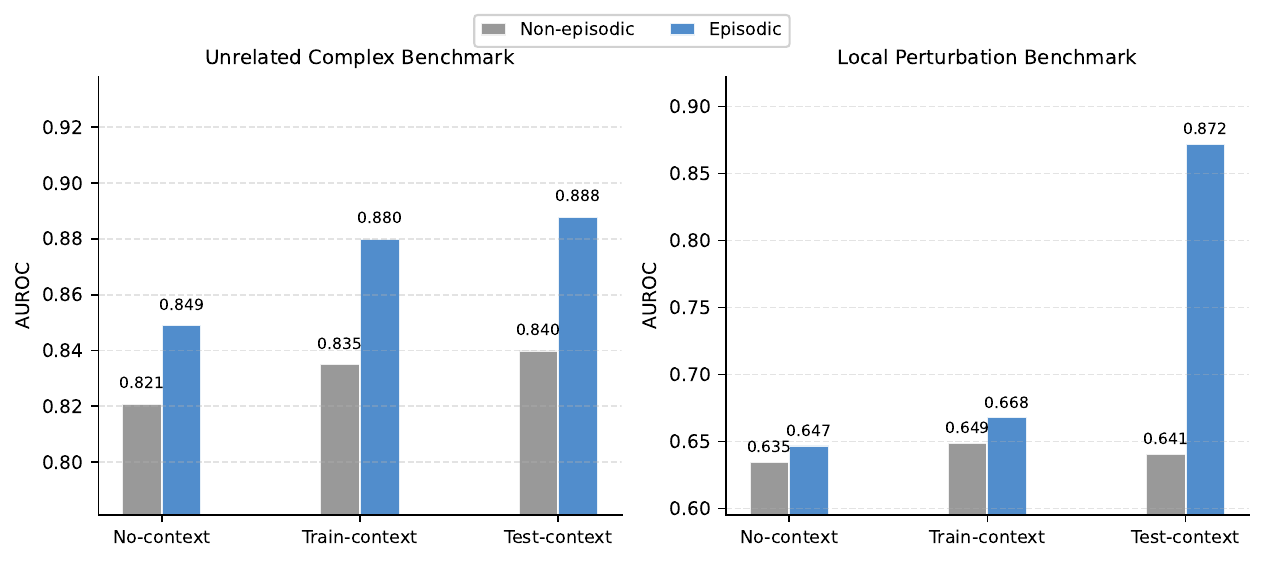}
\caption{Results of the episodic meta-training ablation.
}
\label{fig:episodic_ablation}
\end{figure*}

\subsubsection{\textbf{Test-time demonstrations are more effective than historical demonstrations}}
Across almost all evaluation settings, Test-context consistently outperforms Training-context, indicating that the effectiveness of ICL depends not only on the availability of demonstrations but also on how well they match the target ranking task. While Training-context provides historical affinity comparisons for the target antigen, Test-context draws demonstrations directly from the target inference distribution, offering more faithful evidence of the local affinity landscape. Consequently, the model can perform more effective context-dependent adaptation.

\subsubsection{\textbf{Context benefits fine-grained affinity discrimination}}
The improvements brought by contextual demonstrations are particularly pronounced on the Local Perturbation benchmark, which evaluates the ability to distinguish subtle affinity differences among highly similar antibody variants. In this setting, individual antibody-antigen pairs often contain insufficient information to reliably resolve fine-grained preference differences. By incorporating labeled reference comparisons, contextual demonstrations provide additional evidence for calibrating relative affinity judgments, leading to substantially larger improvements than on the Unrelated Complex benchmark. These results indicate that the benefit of ICL increases as affinity ranking requires finer discrimination.

\subsubsection{\textbf{Context becomes more valuable under distribution shift}}
Compared with the Balanced split, contextual demonstrations yield larger improvements under both the Hard Ab and Hard Ag splits. These two splits intentionally evaluate generalization to previously unseen antibody or antigen families, where the fixed ranking function learned during training is more likely to encounter distribution shift. In this setting, antigen-specific contextual demonstrations provide complementary evidence beyond the global ranking function, leading to larger performance gains. This observation suggests that the value of in-context demonstrations increases as the pretrained ranking function becomes less informative.

Taken together, these observations suggest that the effectiveness of ICL depends on both the relevance of contextual demonstrations and the difficulty of the target ranking task. The largest gains are observed when demonstrations closely match the target inference task and when the model must generalize beyond the training distribution or resolve subtle affinity differences. These findings highlight the complementary role of contextual demonstrations to a fixed global ranking function.


\begin{figure*}[t]
\centering
\includegraphics[width=0.95\linewidth]{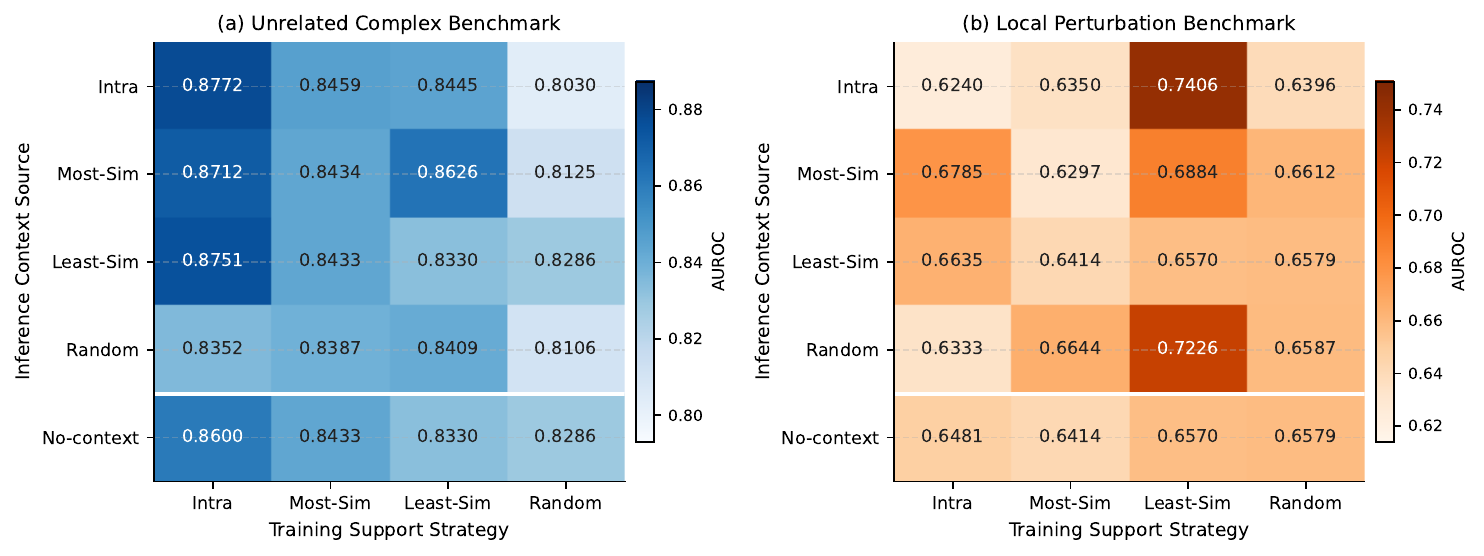}
\caption{AUROC under different training support strategies and inference context sources.
}
\label{fig:support_context_abl}
\end{figure*}
\subsection{Ablation on Episodic Meta-training}

We compare AbICL trained with and without episodic meta-training while keeping the architecture and parameter count identical. The non-episodic variant is trained with the support size fixed to zero throughout training. It can be viewed as the WALLE-Affinity encoder with the same Context Ranking Head, but trained without episodic meta-training.

\subsubsection{\textbf{Episodic meta-training enables effective in-context adaptation}}
As shown in \Cref{fig:episodic_ablation}, contextual demonstrations provide only limited performance gains when the model is trained without episodic meta-training. In contrast, the episodically trained model consistently benefits from contextual demonstrations under both Training-context and Test-context inference, with Test-context yielding the largest performance gains. These results demonstrate that episodic meta-training is essential for learning effective in-context adaptation.

\subsubsection{\textbf{Performance gains arise from episodic learning rather than increased model capacity}}
Under the No-context protocol, the episodically trained model already outperforms its non-episodic counterpart despite having identical architectures and parameter counts. When contextual demonstrations are available, this gap further widens despite both variants using the same Context Ranking Head. This observation indicates that the improvements of AbICL cannot be attributed solely to the additional Context Ranking Head.




\subsection{Ablation on Support and Context Construction}

To investigate how the source of demonstrations affects in-context adaptation, we vary both the training support strategy and the inference context source according to the antigen relationship between demonstrations and query pairs. Specifically, during episodic meta-training, support pairs are constructed using one of four strategies: \textit{Intra-Ag} (from the same antigen), \textit{Most-Sim-Ag} (from the most sequence-similar antigen), \textit{Least-Sim-Ag} (from the least similar detectable antigen identified by MMseqs2), and \textit{Random-Ag} (from a randomly selected different antigen). During inference, we evaluate the \textbf{Training-context} protocol by constructing contextual demonstrations from the training set using the same four strategies, together with a \textit{No-context} setting. Results on the Balanced split are shown in \Cref{fig:support_context_abl}.

\subsubsection{\textbf{Biologically related context remains informative at inference}}

Interestingly, the model remains robust when inference demonstrations are drawn from the most similar or least similar training antigens instead of the same antigen. Although Intra-Ag context generally achieves the best performance, demonstrations from biologically related antigens still provide substantial benefits over random context. This observation suggests that the effectiveness of contextual demonstrations depends not only on exact antigen identity but also on the biological relevance of the provided affinity comparisons.

Additionally, on the Local Perturbation benchmark, we observe a different trend where support strategies involving less similar antigens occasionally outperform Intra-Ag support. One possible explanation is that greater antigen diversity encourages the model to learn more discriminative comparison patterns for resolving subtle affinity differences. We leave a more systematic investigation of this phenomenon to future work.

Overall, these results suggest that contextual demonstrations need not come from the identical antigen to be effective, provided that they remain biologically relevant to the target affinity ranking task.

\section{Limitations}

We evaluate AbICL only on antibody affinity ranking, where contextual demonstrations naturally provide relative preference information. Whether the proposed ICL framework can be extended to affinity regression or affinity change prediction remains an open question and requires further investigation.

Additionally, our ablation study suggests that the construction of contextual demonstrations is more complex than simply selecting the most similar examples.Although Intra-Ag demonstrations generally achieve the best performance, we observe an unexpected trend on the Local Perturbation benchmark, where demonstrations from less similar antigens occasionally lead to better results. This finding indicates that the principles governing effective context construction are not yet fully understood and warrant further study.

\section{Conclusion}
In this work, we introduced AbICL, the first ICL framework for antibody-antigen affinity ranking. Instead of relying solely on a fixed ranking function learned during training, AbICL performs antigen-specific affinity ranking by conditioning on contextual demonstrations through a Context Ranking head trained with episodic meta-training. Extensive experiments on the AbRank benchmark demonstrate that AbICL consistently outperforms existing affinity prediction methods across multiple evaluation settings. Further analyses show that the benefits of ICL become increasingly pronounced under distribution shift and fine-grained affinity discrimination, and are greatest when contextual demonstrations closely match the target inference task. Ablation studies further confirm that these improvements arise from learning effective in-context adaptation rather than simply increasing model capacity. We hope this work provides a new perspective on antibody affinity prediction by treating experimentally characterized affinity comparisons as contextual knowledge available at inference time, and that it encourages broader exploration of ICL for biomolecular prediction tasks.




\newpage
\bibliography{main}
\bibliographystyle{unsrt}

\clearpage
\section*{APPENDIX}

\subsection{Details of Model Structures}
The encoder follows the WALLE-Affinity architecture .
For each antibody-antigen complex, residue-level node features are first extracted using
pre-trained language models: AntiBERTy provides 512-dimensional embeddings
for antibody residues, and ESM-2 provides 1280-dimensional embeddings for
antigen residues.
Each embedding is then processed by a two-layer GCN with hidden dimensions of 128 and 64,
ReLU activations, and edge normalization, operating on a C$\alpha$-distance graph with a
10\,\AA{} cutoff, yielding 64-dimensional node embeddings for both the antibody and antigen.
Graph-level representations are obtained via global mean pooling and concatenated to form
a 128-dimensional complex embedding. Given a pair of complexes $(c_1, c_2)$, the pair representation module produces a 768-dimensional token. The in-context ranking head is a 4-layer Pre-LN Transformer encoder
with a hidden dimension of 768, 8 attention heads, and an FFN dimension of 2048.

\subsection{Details of Support Set Construction}\label{support_build}
For each antigen episode, all ranking pairs of that antigen in the training split are collected. At each iteration, $K$ support pairs are randomly sampled from $\text{Uniform}[0, K_{\max}]$, where $K_{\max} = \min(K_{\text{max}}, N-1)$, $N$ is the total number of pairs for that antigen, and $K_{\text{max}}=8$ by default. The remaining pairs serve as queries, with at least one query guaranteed.

\subsection{Details of Inference Context Construction} \label{context_build}
\subsubsection{Training-context}
Context demonstrations are drawn from training-set pairs of the same antigen as the test antigen, with up to $K_{\max}$ pairs selected. For the Local Perturbation benchmark, where test antigens are point mutants not seen during training, a prefix fallback strategy is applied: if no exact antigen match exists in the training set, the antigen name is progressively truncated (e.g., \texttt{SARS-CoV-2\_Q497K} $\rightarrow$ \texttt{SARS-CoV-2}) until a matching wild-type entry is found.

\subsubsection{Test-context}
Context demonstration and query pairs are both drawn from the test set for the same antigen and are non-overlapping. Specifically, $n_{\text{context}} = \min(K_{\max},, \lfloor N \times r \rfloor)$ pairs are used as context demonstration, where $r$ is the context demonstration ratio, and the remaining pairs serve as queries.

\subsubsection{No-context}
No context demonstrations are provided; predictions rely solely on the encoder representations.

\subsection{Details of Experiments}
\Cref{tab:main-results} presents the main performance comparison on the AbRank benchmark, while \Cref{tab:ablation-episodic} and \Cref{tab:ag-sim-ablation} provide an ablation study on episodic meta-training and a detailed analysis of various training and inference strategies, respectively."
\begin{table}[htbp] 
\centering
\footnotesize 
\setlength{\tabcolsep}{3pt}
\caption{AUROC on the AbRank benchmark.}
\label{tab:main-results}
\begin{tabular}{lccc}
\toprule
\textbf{Model} & \textbf{Balanced} & \textbf{Hard Ab} & \textbf{Hard Ag} \\
\midrule
\multicolumn{4}{c}{\textit{Unrelated Complex Benchmark}} \\
\midrule
AbICL (No-context)      & 0.849 & 0.760 & 0.700 \\
AbICL (Training-context) & 0.880 & 0.838 & 0.711 \\
AbICL (Test-context)     & 0.888 & 0.904 & 0.749 \\
WALLE-Affinity \cite{liu2025abrank}   & 0.866 & 0.763 & 0.746 \\
ESM-2 + AntiBERTy \cite{ruffolo2021deciphering} & 0.761 & 0.719 & 0.758 \\
Mint \cite{ullanat2026learning}             & 0.775 & 0.741 & 0.688 \\
\midrule
\multicolumn{4}{c}{\textit{Local Perturbation Benchmark}} \\
\midrule
AbICL (No-context)      & 0.647 & 0.490 & 0.572 \\
AbICL (Training-context) & 0.668 & 0.588 & 0.581 \\
AbICL (Test-context)     & 0.872 & 0.787 & 0.778 \\
WALLE-Affinity \cite{liu2025abrank}   & 0.671 & 0.581 & 0.637 \\
ESM-2 + AntiBERTy \cite{ruffolo2021deciphering} & 0.595 & 0.516 & 0.532 \\
Mint \cite{ullanat2026learning}             & 0.625 & 0.538 & 0.497 \\
\bottomrule
\end{tabular}
\end{table}
\begin{table}[htbp]
\centering
\footnotesize %
\setlength{\tabcolsep}{4pt} %
\caption{Results of the episodic meta-training ablation.}
\label{tab:ablation-episodic}

\begin{tabular}{llcc}
\toprule
\textbf{Model} & \textbf{Support Context} & \textbf{Gen.} & \textbf{Pert.} \\ %
\midrule
\multirow{3}{*}{AbICL (non-epi.)} 
    & No-context     & 0.821 & 0.635 \\
    & Training-context  & 0.835 & 0.649 \\
    & Test-context   & 0.840 & 0.641 \\
\midrule
\multirow{3}{*}{AbICL (episodic)}
    & No-context     & 0.849 & 0.647 \\
    & Training-context  & 0.880 & 0.668 \\
    & Test-context   & 0.888 & 0.872 \\
\bottomrule
\multicolumn{4}{l}{\scriptsize *Gen.: Generalization, Pert.: Perturbation.} %
\end{tabular}
\end{table}

\begin{table}[htbp]
\centering
\footnotesize
\setlength{\tabcolsep}{6pt} %
\caption{AUROC under different training support strategies and inference context sources.}
\label{tab:ag-sim-ablation}

\begin{tabular}{lcccc}
\toprule
& \multicolumn{4}{c}{\textbf{Training Strategy}} \\
\cmidrule(lr){2-5}
\textbf{Inf. Context} & \textbf{Intra} & \textbf{M-Sim} & \textbf{L-Sim} & \textbf{Rand} \\
\midrule
\multicolumn{5}{l}{\textit{(a) Generalization Benchmark}} \\ 
\midrule
Intra       & 0.8772 & 0.8459 & 0.8445 & 0.8030 \\
M-Sim       & 0.8712 & 0.8434 & 0.8626 & 0.8125 \\
L-Sim       & 0.8751 & 0.8433 & 0.8330 & 0.8286 \\
Rand        & 0.8352 & 0.8387 & 0.8409 & 0.8106 \\
No-context  & 0.8600 & 0.8433 & 0.8330 & 0.8286 \\
\midrule
\multicolumn{5}{l}{\textit{(b) Local Perturbation Benchmark}} \\
\midrule
Intra       & 0.6240 & 0.6350 & 0.7406 & 0.6396 \\
M-Sim       & 0.6785 & 0.6297 & 0.6884 & 0.6612 \\
L-Sim       & 0.6635 & 0.6414 & 0.6570 & 0.6579 \\
Rand        & 0.6333 & 0.6644 & 0.7226 & 0.6587 \\
No-context  & 0.6481 & 0.6414 & 0.6570 & 0.6579 \\
\bottomrule
\end{tabular}
\end{table}

\end{document}